\documentclass[10pt,twocolumn,letterpaper]{article}

\usepackage{cvpr}              

\usepackage{graphicx}
\usepackage{amsmath}
\usepackage{amssymb}
\usepackage{booktabs}
\usepackage{lipsum,graphicx}
\usepackage{cuted}
\usepackage{subcaption}

\newcommand{\cI}{\mathcal{I}}

\newcommand{\cT}{\mathcal{T}}

\usepackage{pifont}
\usepackage{xcolor}
\usepackage{color, colortbl}
\newcommand{\cmark}{\textcolor{green!80!black}{\ding{51}}}
\newcommand{\xmark}{\textcolor{red}{\ding{55}}}
\definecolor{mygray}{rgb}{0.95,0.95,0.95}
\definecolor{mypink}{rgb}{1,0.49,0.51}
\definecolor{myorange}{rgb}{1,0.75,0.40}
\usepackage[accsupp]{axessibility}  

\usepackage[pagebackref,breaklinks,colorlinks]{hyperref}

\usepackage[capitalize]{cleveref}
\crefname{section}{Sec.}{Secs.}
\Crefname{section}{Section}{Sections}
\Crefname{table}{Table}{Tables}
\crefname{table}{Tab.}{Tabs.}

\def\confName{CVPR}
\def\confYear{2023}

\begin{document}

\title{TempSAL - Uncovering Temporal Information for Deep Saliency Prediction}
  
\author{Bahar Aydemir, Ludo Hoffstetter, Tong Zhang, Mathieu Salzmann, Sabine Süsstrunk \\
School of Computer and Communication Sciences, EPFL, Switzerland\\
{\tt\small \{bahar.aydemir, tong.zhang, mathieu.salzmann, sabine.susstrunk\}@epfl.ch \vspace*{-20pt}}
}
\maketitle
\begin{strip}
\includegraphics[width=1\textwidth]{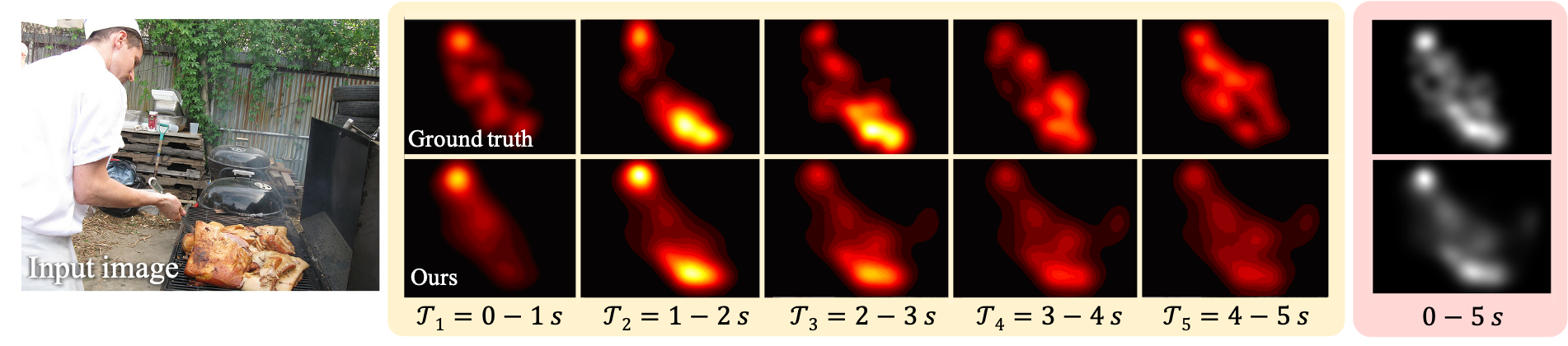}
\vspace*{-17pt}
\captionof{figure}{An example of how human attention evolves over time. Top row: \textbf{\textcolor{myorange}{Temporal}} and \textbf{\textcolor{mypink}{image}} saliency ground truth from the SALICON dataset~\cite{salicon}. Bottom row: Our \textbf{\textcolor{myorange}{temporal}} and \textbf{\textcolor{mypink}{image}}  saliency predictions. Each temporal saliency map $\mathcal{T}_i$, $i \in \{1,\ldots,5\}$ represents one second of observation time. Note that in $\mathcal{T}_1$, the chef is salient, while in  $\mathcal{T}_2$ and  $\mathcal{T}_3$, the food on the barbecue becomes the most salient region in this scene. We can predict the temporal saliency maps for each interval separately, or combine them to create a single, refined image saliency map for the entire observation period.  
\label{fig:teaser}}

\end{strip}

\begin{abstract}
Deep saliency prediction algorithms complement the object recognition features, they typically rely on additional information, such as scene context, semantic relationships, gaze direction, and object dissimilarity. However, none of these models consider the temporal nature of gaze shifts during image observation. We introduce a novel saliency prediction model that learns to output saliency maps in sequential time intervals by exploiting human temporal attention patterns.
Our approach locally modulates the saliency predictions by combining the learned temporal maps. Our experiments show that our method outperforms the state-of-the-art models, including a multi-duration saliency model, on the SALICON benchmark and CodeCharts1k dataset. Our code will
be publicly available on GitHub\footnote{\href{https://ivrl.github.io/Tempsal//}{\textcolor{magenta}{\tt https://ivrl.github.io/Tempsal/}}}.


\end{abstract}

\section{Introduction}

Humans have developed attention mechanisms that allow them to selectively focus on the important parts of a scene. Saliency prediction algorithms aim to computationally detect these regions that stand out relative to their surroundings. These predictions have numerous applications in image compression~\cite{saliencycompression}, image enhancement~\cite{ZHAO20144039}, image retargeting~\cite{AC}, rendering~\cite{steinberger2012ray}, and segmentation~\cite{li2011saliency}.

Since the seminal work of Itti et al.~\cite{Itti98}, 
many have developed solutions using both handcrafted features~\cite{salmap_handcrafted} and deep ones~\cite{vigsal,mlnet,saliconmodel,deepgaze2,dsclrcn,dinet}. Nowadays, employing deep neural networks is preferred in saliency prediction as they outperform bottom-up models. These methods typically depend on pretrained object recognition networks to extract features from the input image~\cite{linardos}. In addition to these features, scene context~\cite{wang2020salient}, object co-occurrence~\cite{external-knowledge-saliency}, and dissimilarity~\cite{objdissim} have been exploited to improve the saliency prediction. However, while these approaches model the scene context and objects, they fail to consider that humans dynamically observe scenes~\cite{Yarbus1967}. In neuroscience, the inhibition of return paradigm states that a suppression mechanism reduces visual attention towards recently attended objects\cite{ior} and  encourages selective attention to novel regions. Motivated by this principle, we develop a saliency prediction model that  incorporates temporal information. 

Fosco et al.~\cite{fosco2020howmuch} also exploit temporal information in saliency prediction, but they consider snapshots containing observations up to 0.5, 3, and 5 seconds, thus not leveraging saliency trajectory but rather saliency accumulation. By contrast, here, we model consecutive time slices, connecting our approach more directly with the human gaze and thus opening the door to automated visual appeal assessment in applications such as website design~\cite{website50ms}, advertisement~\cite{ads} and infographics~\cite{infographics}.

To achieve this, we show that when viewing images, human attention yields temporally evolving patterns, and we introduce a network capable of exploiting
this temporal information for saliency prediction.
Specifically, our model learns time-specific predictions and is  able to combine them with a conventional image saliency map to obtain a temporally modulated image saliency prediction.

\par We use our method on SALICON~\cite{salicon} and CodeCharts1k~\cite{fosco2020howmuch} datasets which contain temporal information, unlike the other popular saliency datasets such as CAT2000~\cite{cat2000} and MIT1003~\cite{Judd_2009}. We show the benefits of estimating temporal saliency to encourage the community to publish the temporal information of their data along with the final saliency maps. Note that, existing works collect saliency data by conducting psychophysical experiments \cite{cat2000,salicon,Judd_2009,fosco2020howmuch}. The attention data recorded during these experiments already includes temporal information. Therefore, no further experiments are required. 

As evidenced by our experiments, using temporal information consistently boosts the accuracy of the baseline network, enabling us to consistently outperform the state-of-the-art models on the SALICON saliency benchmark.  Moreover,  in the CodeCharts1k dataset we outperform this multi-duration model in two out of three metrics. 

We summarize our contributions as follows:
\begin{itemize}
\item We evidence the presence of temporally evolving patterns in human attention.
\item We show that temporal information in the form of a saliency trajectory is important for saliency prediction in natural images, providing an investigation of the SALICON dataset for temporal attention shifts.
\item We introduce a novel, saliency prediction model, namely TempSAL, capable of simultaneously predicting conventional image saliency and temporal saliency trajectories.
\item We propose a spatiotemporal mixing module that learns time dependent patterns from temporal saliency maps. Our approach outperforms the state-of-the-art image saliency models that either do not consider temporal information or encode it in a cumulative manner.

\end{itemize}

\section{Related work}


\subsection{Saliency prediction for natural images}
Early saliency prediction methods were biologically inspired and bottom up. In particular, Itti et al. used color, intensity, and orientation contrast \cite{Itti98}. Goferman employed global and local contrast as contextual cues \cite{goferman2011context}. Judd et al. \cite{Judd_2009} further incorporated mid-level and high-level semantic features, using horizon, face, person, and car detectors. Later, Vig et al. \cite{vigsal} showed that deep neural networks can be applied to saliency prediction. Yet, saliency prediction lacks the large scale annotated datasets that are available for image classification tasks\cite{deng2009imagenet}, which prevents training robust models. To overcome this, Kummerer et al. \cite{kummerer2015deep} showed that using pretrained object recognition networks significantly improves saliency predictions. Subsequent state-of-the-art models such as EML-Net \cite{eml}, DeepGaze2 \cite{deepgaze2}, and SALICON \cite{saliconmodel} similarly use pretrained convolutional neural network (VGG\cite{vgg}) encoders. 
Recent works utilize additional sources of information such as scene context \cite{KRONER2020261}, external knowledge \cite{external-knowledge-saliency} and object dissimilarity \cite{objdissim} to improve saliency prediction. Yet, none of these methods take into account the temporal evolution of human gaze, which occurs even when the image stimuli are static. In our work, we make use of these temporal patterns as an additional source of information to boost conventional image saliency prediction. 


\subsection{Multi-duration saliency}
Existing image saliency ground-truth maps include all fixations made throughout the observation period. Aggregating all these fixations that have different timestamps into a single ground truth map results in the loss of temporal information. Representing the fixations as scanpaths retains the temporal clues by encoding the change of gaze of an individual over time. However, merging numerous scanpaths is challenging \cite{scanpaths}. Fosco et al. \cite{fosco2020howmuch} proposed \textit{multi-duration saliency} to characterize the attention of a group of individuals while taking into account time-dependent attention shifts. 
The temporal maps they rely on, however, encode the attention distribution of many observers across overlapping time periods of increasing durations. 
While this is a convenient way of capturing a population's attention patterns, it does not reflect the saliency trajectory over time.
 Similarly, \cite{mishra} use order of fixations as a sequential metadata for deep supervision but they do not model evolution of attention through time.
In our work, we model multi-duration saliency to analyze underlying attention patterns by using \emph{mutually exclusive} time slices. We provide this temporal information to our spatiotemporal mixing module to refine the initial image saliency prediction with temporal information. Moreover, this lets us predict temporal saliency maps for each second of attention. 

\section{Temporal saliency data analysis}

\subsection{Datasets}
SALICON \cite{salicon} is the largest human attention dataset on natural images. It was created via a crowdsourced mouse tracking experiment, which was shown to be similar to eye-tracking \cite{salicon} and widely used in the saliency prediction literature. SALICON consists of 10000 training, 5000 validation and 5000 test images from the MS-COCO dataset \cite{mscoco}. The SALICON dataset provides saliency maps, fixations, and gaze points for each image and observer. A gaze point is a raw data point recorded by a tracking device. It describes the spatial coordinates of the eye/mouse on the associated stimuli at a given timestamp. Conversely, fixations describe the coordinates of the long pause when the eyes are fixated on an image detail. Following common practice in eye tracking experiments, Jiang et al. \cite{salicon} grouped spatially and temporally close gaze points to create fixations. Since the fixations were created by grouping multiple gaze points, they do not have an associated timestamp. 
To address this, SALICON-MD \cite{fosco2020howmuch} assumes that the fixations are uniformly distributed across the total viewing time. We provide a finer approximation for recovering the fixations' timestamps, by minimizing the spatial and temporal distance between a fixation and the nearest gaze point. We refer the reader to the supplementary material for the details of this approximation process. 
\par We also assess the performance and generalization ability of our method using CodeCharts1k \cite{fosco2020howmuch}, which is the first saliency dataset to report multiple viewing durations. The dataset consists of 1000 images and saliency maps collected via crowdsourcing which correspond to 0.5, 3, and 5 seconds of viewing. We present the results in Section \ref{sec:multiresults}.

\subsection{Temporal patterns in the dataset}
\label{sec:temporal}
In this section, we examine how temporality evolves during human visual attention.
To observe the evolution of attention over time, we slice the data into five slices, one for each second of observation. In particular, we inspect the dissimilarity between slices, the agreement with the average, and the distribution of fixations in the time-saliency space. 
\begin{figure}[h]
\vspace{-8pt}
    \centering
    \includegraphics[width=0.475\textwidth]{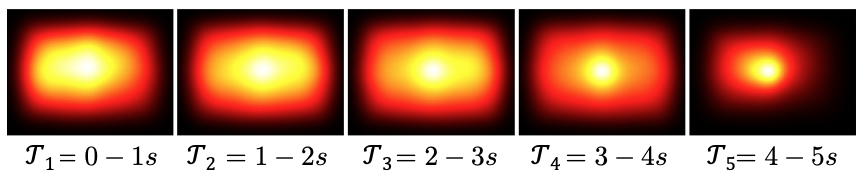}
    \vspace{-20pt}
    \caption{Average heat maps for each one second interval. Note that a center-bias occurs, similar to image saliency prediction's average ground-truth maps.}
    \label{fig:avg-slices}
\end{figure}
\vspace{-5pt}%
\par 
\textbf{Average maps.} Viewing patterns in image saliency experiments tend to show a concentration towards the image center \cite{centerbias}, which is known as the photographer's bias or center bias. We observe a similar spatial bias for each temporal slice, shown in Figure~\ref{fig:avg-slices}, where we plot the average heat maps for each temporal slice. Note that the gaze tends to converge to the center of the image as time passes. This means that the observers revisit the previously seen important center regions \cite{Yarbus1967}.

\par
\begin{figure}[h]
\centering
\includegraphics[width=0.48\textwidth]{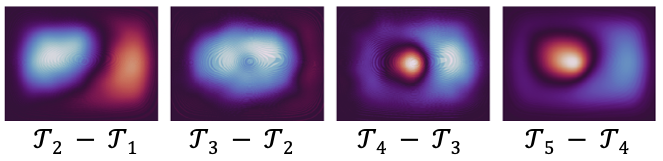}
\caption{Differences of the consecutive average temporal slices shown in Fig.~\ref{fig:avg-slices}. Red indicates regions of increased attention whereas blue indicates decreased attention.  }
\label{fig:avg-shifts}
\end{figure}

 
We plot the differences of the consecutive average temporal slices in Figure~\ref{fig:avg-shifts} to illustrate attention shifts. Light blue indicates the regions with reduced attention, whereas (light) red indicates increased attention. 
We observe that attention shifts from left to right, with a subsequent dispersion from the center towards the corners. Then, attention increases at the center of the image, slightly skewed to the left. Interestingly, the trend (especially in $\mathcal{T}_{2} - \mathcal{T}_{1}$) coincides with the western left-to-right reading direction \cite{eye-culture}. 

\begin{table}

\setlength{\tabcolsep}{8pt}
\renewcommand{\arraystretch}{1}
\begin{tabular}{cccccc}
 & $\mathcal{T}_1$& $\mathcal{T}_2$  & $\mathcal{T}_3$ & $\mathcal{T}_4$ & $\mathcal{T}_5$  \\
\cmidrule(lr){2-2}\cmidrule(lr){3-3}\cmidrule(lr){4-4}\cmidrule(lr){5-5}\cmidrule(lr){6-6}
 $\mathcal{T}_1$& $\bold{1.00}$ & $\underline{0.70}$ & $0.54$ & $0.50$ & $0.54$ \\
$\mathcal{T}_2$ & $0.70$ & $\bold{1.00}$ & $\underline{0.73}$ & $0.66$ & $0.65$ \\
$\mathcal{T}_3$ & $0.54$ & $0.73$ & $\bold{1.00}$ & $\underline{0.75}$ & $0.70$ \\
$\mathcal{T}_4$ & $0.50$ & $0.66$ & $\underline{0.75}$ & $\bold{1.00}$ & $0.73$ \\
$\mathcal{T}_5$ & $0.54$ & $0.65$ & $0.70$ & $\underline{0.73}$ & $\bold{1.00}$ \\
	
\rowcolor{mygray}
Average &  $0.66$ &	 $0.75$ &  	$0.74$ &  	$0.73$ &	  $0.72$   \\
\end{tabular}
\caption{Correlation scores (CC) of the temporal slices with each other in a single image, averaged over all images. All slices show more similarity to their direct temporal neighbors. The last row shows the average similarity of a slice with the other slices, $\mathcal{T}_1$ being the most dissimilar one.}
\vspace{-10pt}
\label{tab:corr-table}
\end{table}
\textbf{Inter-slice similarity across time.}
We expect temporal saliency slices to be more similar to their closer-in-time slices than to the ones further away since human attention is continuous over time. Table~\ref{tab:corr-table} contains
the correlation coefficients between each pair of saliency slices in a single image, averaged over all images. We calculate the correlation coefficient between slices $\mathcal{T}_{j}$ and $\mathcal{T}_{k}$ as
\begin{equation}
    \mathrm{CC}(\mathcal{T}_j,\mathcal{T}_k) = \frac{1}{N}\sum_{n=i}^{N} \mathrm{CC}(\mathcal{T}_{ij},\mathcal{T}_{ik}) ,\quad j,k \in \{1,\ldots,5\},
\end{equation}
where $N$ is the total number of images, and $\mathcal{T}_{ij}$ and $\mathcal{T}_{ik}$ denote the $j^{th}$ and $k^{th}$ slice of the $i^{th}$ image.

By calculating t-test scores on the pairwise comparisons, we observe that all of the pairwise differences except $\mathcal{T}_1,\mathcal{T}_3$ and $\mathcal{T}_1,\mathcal{T}_5$ are statistically significant ($p<0.01$). Thus, the attention residuals between different time intervals in one image are significantly different. We provide more details in the supplementary material. \\

\textbf{Intra-slice similarity across images.}
We also investigate the deviation of each slice from its respective average slices. The average slices are depicted in Figure~\ref{fig:avg-slices}. 
Table~\ref{tab:cc-perimg} shows the deviation of a slice from the average time slices per image. We compute CC scores between a single slice and the corresponding average slice as 
\begin{equation}
    \mathrm{CC}(\mathcal{T}_j,A_j) = \frac{1}{N}\sum_{n=i}^{N} \mathrm{CC}(\mathcal{T}_{ij},A_{j}) ,\quad j \in \{1,\ldots,5\},
\end{equation}
where 
$A_{j}$ denotes the $j^{th}$ average slice.

We average the scores across all images.
Higher values of CC indicate more agreement with the average whereas lower values of CC indicate more deviation from the average. Note that the similarity with the average across images decreases with time, except for the last slice. This can be explained by the more prominent center bias in $\mathcal{T}_5-\mathcal{T}_4$ as seen in Figure~\ref{fig:avg-slices}. $\mathcal{T}_1$ has the least deviation from the average
by a significant margin $(p\ll0.01)$. This shows that humans tend to look at similar places at first, and then their attention scatters around to less important regions.
 \\

\begin{figure}[h]
\centering
\includegraphics[width=0.5\textwidth]{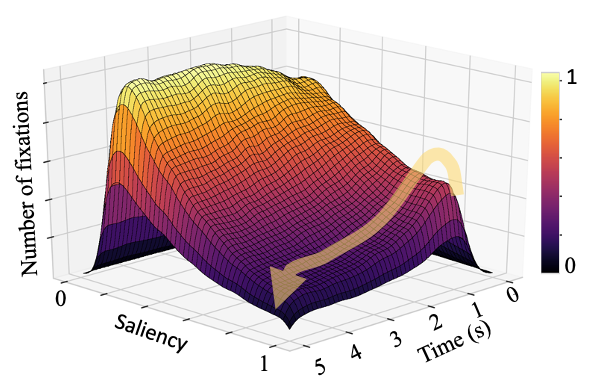}
\caption{ Number of fixations with their respective saliency values and timestamps. Lighter colors indicate higher number of occurrences while darker areas denote fewer occurrences. We see that late fixations tend to be less salient, which can be seen as the decrease in the number of salient fixations along the arrow.  The most salient fixations appear at approximately 1s. }
\label{fig:fixhist}
\end{figure}

\textbf{Early and late fixations versus saliency.}
Lastly, we investigate the relationship between the fixation timestamps and their respective saliency values. We assign a saliency value to each fixation as the normalized pixel value in the corresponding saliency map. We plot the histogram of number of fixations with their saliency and timestamp values in Figure~\ref{fig:fixhist}. The fixation time stamps range from 0 to 5000 ms and the saliency values range from 0 to 1. Late fixations tend to have lower saliency values than earlier fixations, as indicated by the darker color towards the bottom right corner. That is, the first region we glance at in an image is more important (salient) than the following regions~\cite{Itti98,salmap_handcrafted}.

\begin{table}

\setlength{\tabcolsep}{8pt}
\renewcommand{\arraystretch}{1}
\begin{tabular}{lccccc}
 & $\mathcal{T}_1$& $\mathcal{T}_2$& $\mathcal{T}_3$& $\mathcal{T}_4$ & $\mathcal{T}_5$ \\
 \cmidrule(lr){2-2}\cmidrule(lr){3-3}\cmidrule(lr){4-4}\cmidrule(lr){5-5}\cmidrule(lr){6-6}
 CC & $0.574$ & $0.433$ & $0.431$ & $0.426$ & $0.447$ \\

\end{tabular}

\caption{Correlation scores (CC) of each time slice in a single image with the average maps presented in Figure~\ref{fig:avg-slices}, averaged over all images. The similarity between slices across images decreases with time, with the exception of the last slice.}
\label{tab:cc-perimg}
\vspace{-8pt}
\end{table}
 
\section{Methodology}
\begin{figure*}[t]
\centering
\includegraphics[width=1\textwidth]{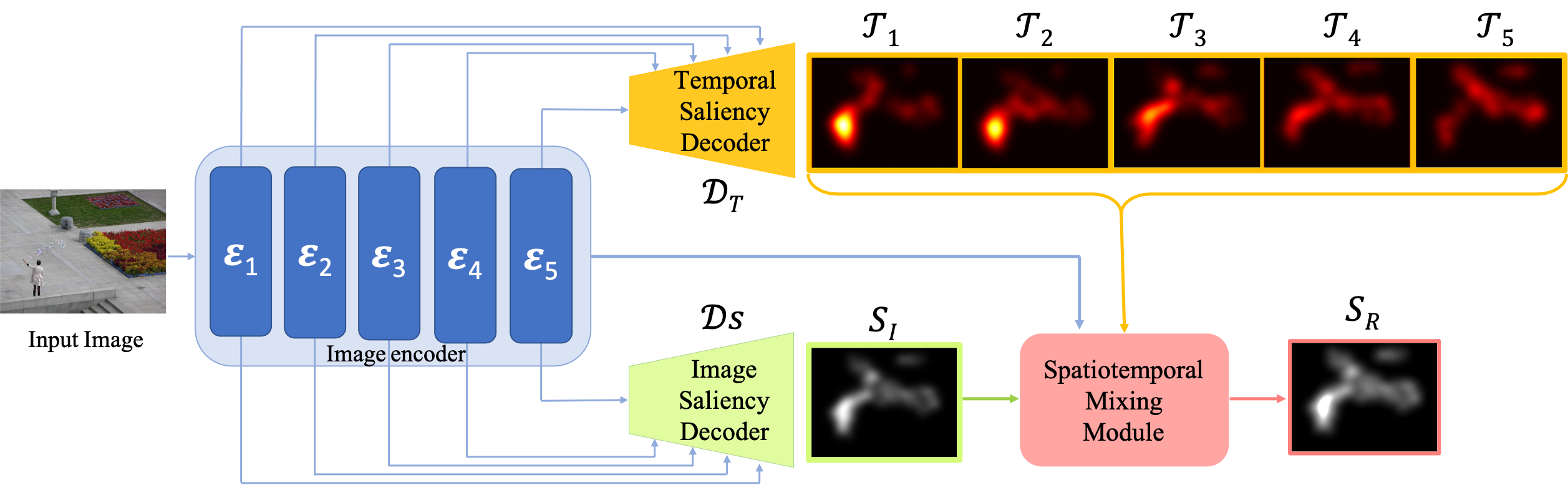}
 
\caption{\textbf{Overview of the proposed architecture.} We encode image features into encoder blocks consisting of multi-level image features. We then pass these blocks to the temporal saliency decoder (shown in \textbf{\textcolor{myorange}{orange}}) to decode them into temporal saliency predictions, which are saliency maps in sequential time intervals. In parallel, the image saliency decoder (shown in \textbf{\textcolor{green}{green}}) decodes the encoder blocks into an image saliency prediction. We then combine (1) the temporal saliency maps, (2) the image saliency map, and (3) the encoder blocks in the spatiotemporal mixing module (shown in \textbf{\textcolor{mypink}{pink}}).  (Best viewed in color.)  }
\label{fig:mainmodel}
\end{figure*}

\subsection{Temporal slices}

We aim to recover fixation timestamps to train models with this temporal information. We extract temporal slices by grouping the fixations in several time-intervals and, following common practice \cite{AUCJ}, blurring with a Gaussian kernel. We break down the fixations into time slices with two time-slicing (grouping) alternatives, namely equal duration and equal distribution. The equal duration model outperforms the equal distribution one and is easier to interpret; we present a comparison of these two alternatives as an ablation study in the supplementary material. 

\subsection{Temporal saliency model}
Let us now introduce our framework that exploits temporal human attention information. Our model is depicted in Figure~\ref{fig:mainmodel}. We extract image features using a pre-trained object recognition encoder \cite{pnasnet}. Then, we decode these features by a temporal slice decoder to obtain one saliency map per time slice. These temporal saliency slices are useful in automated visual appeal assessment in applications such as website design~\cite{website50ms}, advertisement~\cite{ads} and infographics~\cite{infographics} In parallel, we decode the same image features into an initial image saliency prediction. Finally, we combine the temporal slices and the image saliency predictions in the spatiotemporal mixing module to produce a final image saliency map.
We describe each component in detail in the following sections.

\subsubsection{Image encoder and saliency decoders}
Following the previous saliency prediction architectures \cite{deepgaze2, linardos, simplenet},  we first encode the input image with a pre-trained image classification network, in our case PNASNet-5 \cite{pnasnet}. We extract encoded features at various levels for multi-level integration, similar to a U-Net structure \cite{unet}. Formally, we denote the image encoder as
\begin{equation}
    \mathcal{E(I)} = [\mathcal{E}_{i}],\quad i \in \{1,\ldots,5\},
\end{equation}
where $\mathcal{I}$ is the input image, and $\mathcal{E}_{i}$ the $i^{\text{th}}$ encoder block. The output of $\mathcal{E}(\cdot)$ therefore is a 5D vector. The early encoder blocks extract low-level features, such as edges, color, and contrast, while the later blocks encode high-level semantics. We pass these blocks to the temporal saliency decoder, the image saliency decoder, and the spatiotemporal mixing module.

Our temporal slice decoder, namely  $\mathcal{D}_{T}$, processes the encoder blocks with four 3x3 convolution layers followed by ReLU functions, integrating one encoder block after each convolution. Later, two 3x3 convolution layers with a ReLU in-between and a sigmoid function at the end produce $n$ temporal saliency maps. Formally, we write the temporal saliency decoder as
\begin{equation}
    \mathcal{D}_{T}\big(\mathcal{E}(\cI)\big) = [\mathcal{T}_{n}] =: \cT,\quad n \in \{1,\ldots,5\}
\end{equation}
where $\mathcal{T}_{n}$ denotes the $n^{\text{th}}$ temporal saliency slice. Through this branch of the network, a single image input produces $n$ temporal saliency slices. We use this component to provide temporal predictions to the spatiotemporal mixing module.

Our image saliency decoder, namely  $\mathcal{D}_{S}$, has the same structure as $\mathcal{D}_{T}$, with the exception of the number of output channels. This component produces a single map $\mathcal{S}_{I}$, which corresponds to the conventional image saliency map of the input image. As such, we can write
\begin{equation}
   \mathcal{S}_{I} = \mathcal{D}_{S}\big(\mathcal{E}(\cI)\big).
\end{equation}
We use this module to provide cumulative saliency information to the spatiotemporal mixing module. 

\subsubsection{Spatiotemporal Mixing Module}

To incorporate temporal information into the saliency prediction, we introduce a module that combines temporal and spatial saliency maps.
Our design is inspired by the
feature pyramid networks~\cite{lin2017feature}, which has the benefit of integrating features from multiple levels of the encoder,
thus capturing low-level cues (such as color contrast, brightness,
and edges) and high-level ones (such as semantics and scene context). Such cues have been shown to be critical for saliency estimation ~\cite{NSS,goferman2011context,KRONER2020261}.
This module takes temporal saliency predictions, the initial image saliency prediction, and the encoded image feature blocks as input. We write this as
\begin{equation}
    \mathcal{S}_{R} = \mathrm{SMM}\big(\mathcal{E}(\cI),\mathcal{T},\mathcal{S}_{I} \big)  ,\ n \in \{1,\ldots,5\},
\end{equation}
where $\mathcal{S}_{R}$ denotes the final, refined image saliency map.

\begin{figure}[t]
\centering

 \includegraphics[width=0.5\textwidth]{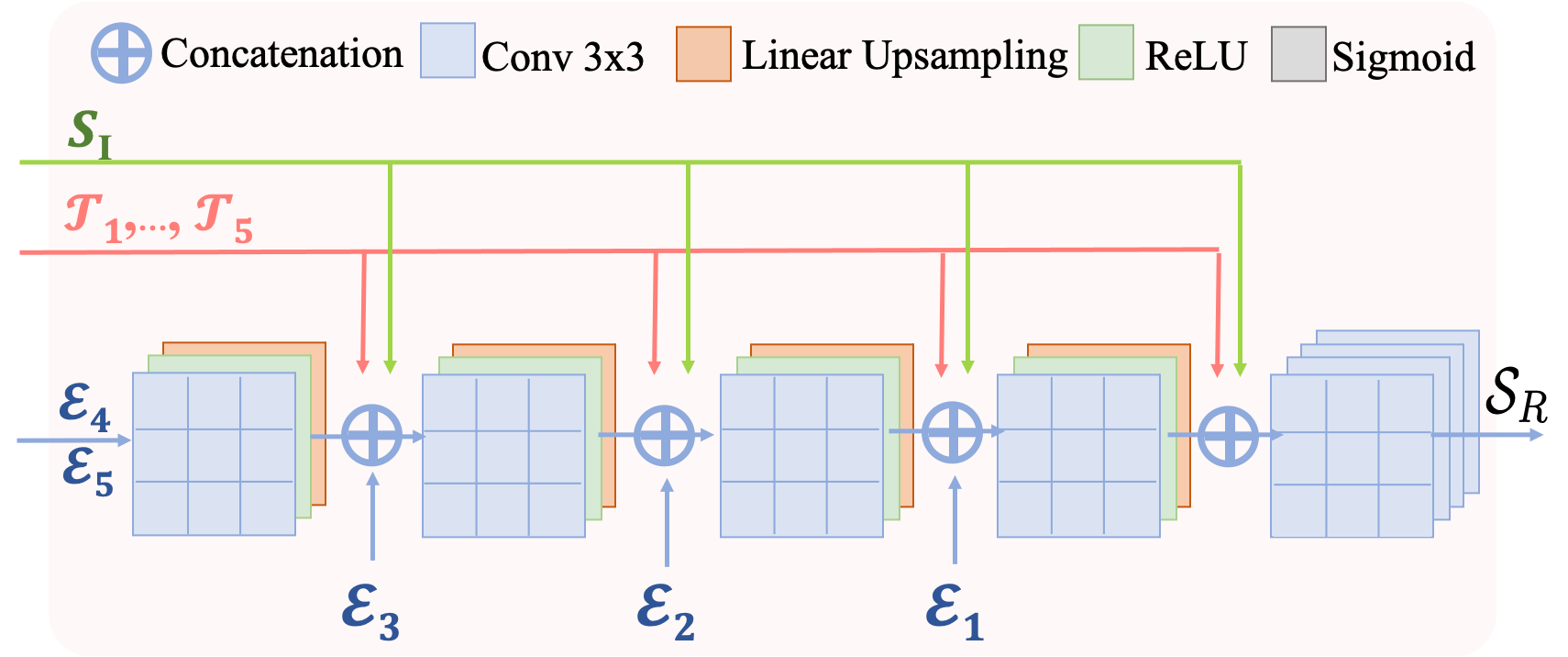}

\caption{The spatiotemporal mixing module combines temporal saliency predictions with the conventional image saliency prediction with multi-level image feature integration. $\mathcal{S}_{I}$ denotes the predicted image saliency map, $\mathcal{T}
_{1,..,5}$ the temporal saliency predictions for $n$ frames, and $\mathcal{E}_{1,..,5}$ the encoder blocks.  This multi-level integration scheme provides information from earlier layers of the network to the next blocks in this module. $\mathcal{S}_{R}$ denotes the temporally refined image saliency map output. }
\label{fig:mixing}

\end{figure}

The module architecture is shown in Figure~\ref{fig:mixing}. It takes the last two encoder blocks $[\mathcal{E}_{5}, \mathcal{E}_{4}]$ and passes them through a 3x3 convolution. We then concatenate the other encoder blocks with the image saliency and temporal saliency maps passing through 3x3 convolution, ReLU, and linear upsampling to keep the spatial dimensions consistent. We add the static
and temporal saliency maps to the encoded image features
at each block to prevent the information in these maps from
vanishing. 
In the last step, we only add the saliency maps to output a final refined saliency map $\mathcal{S}_{R}$. The final design of the module (i.e., combining E4 and E5) is based on empirical performance.
This module eliminates the need for optimizing a weight parameter between the spatial and temporal maps. It can also modulate the maps within the spatial range of convolutions, which allows the selection of different regions from different maps. 

\subsubsection{Loss Functions}
To train our network, 
we use the Kullback- Leibler divergence (KL) \cite{kld} and the Correlation Coefficient (CC) \cite{ccmetric} between the predicted and ground-truth saliency maps. First, we train the temporal branch using
\begin{equation}
    \mathcal{L}_{1}(\cI) = \lambda_{1} * \mathrm{CC}(GT_{n},\mathcal{T}_{n}) + \beta_{1} * \mathrm{KL}(GT_{n},\mathcal{T}_{n}),
\end{equation}
where $GT_{n}$ denotes the temporal ground truth for the $n^{th}$ slice. We then freeze the weights in this component and train the spatiotemporal mixing module using
\begin{equation}
    \mathcal{L}_{2}(\cI) = \lambda_{2} * \mathrm{CC}(GT,\mathcal{S}_{R}) + \beta_{2} * \mathrm{KL}(GT,\mathcal{S}_{R}),
\end{equation}
where $GT$ is the image saliency ground truth for image $I$.

\section{Experiments and Results}
\subsection{Experimental Setup }
We use a batch size of 32 and an initial learning rate of 1e-4, reduced by a factor of ten every two epochs. We train the temporal branch first and then freeze the weights. We found that 10 epochs of training on SALICON was sufficient. For SALICON, we used the provided test, train, and validation splits. 
\subsection{Metrics}
We evaluate the obtained saliency predictions according to the following standard metrics used by the community.\\
\textbf{Area Under the Curve (AUC) \cite{AUCJ}}: Saliency prediction can be interpreted as classifying fixation vs non-fixation points. The area under the ROC curve shows the trade-off between true positives (TP) and false positives (FP). A higher AUC score indicates less FPs.  While AUC computes the
TP and FP rates using all the ground-truth fixation points, \textbf{sAUC}\cite{sAUC} samples FP points from ground-truth fixations of other observers, compensating for center bias in natural images by taking for inter- and intra-observer variability into account. \\
\textbf{Normalized Scanpath Saliency (NSS)} \cite{NSS}: This metric compares the predicted saliency values at the ground-truth fixation points to the average predicted saliency. An NSS score of one indicates that the predicted saliency values at the ground-truth fixation points are one standard deviation above the average.\\
\textbf{Kullback - Leibler Divergence (KL)} \cite{kld}: The KL measures the cumulative  distance between the predicted and the ground-truth saliency maps. A KL score close to zero indicates a better approximation of the ground-truth saliency map by the predicted one.\\
\textbf{Pearson’s correlation coefficient (CC)} \cite{ccmetric}: This metric measures the linear relationship between the predicted and ground-truth saliency maps. It ranges from -1 to 1. A CC score close to one indicates a strong linear correlation between the two maps.\\
\textbf{Similarity (SIM) score \cite{simmetric}:} The similarity score sums the minimum value between the predicted and the ground-truth saliency maps over all pixels. A similarity score of 1 indicates a perfect prediction since both of the maps are probability distributions summing to 1.\\
\textbf{Information Gain (IG) score \cite{infogain}:} The information gain is a information-theoretic metric which measures the difference in average log-likelihood between the predicted saliency map and center-bias prior.

\subsection{Quantitative Results}\mbox{}
We compare our method with the state-of-the-art models, namely SAM-Resnet \cite{sam}, MSI-Net, GazeGAN, MDNSal \cite{simplenet}, SimpleNet \cite{simplenet}, DeepGaze IIE \cite{linardos}, and MD-SEM \cite{fosco2020howmuch}, in image saliency prediction. Our model outperforms these methods in five out of seven metrics, showing the benefit of incorporating temporal information. Moreover, our model outperforms the only other multi-duration saliency model in image saliency prediction by a significant margin. Furthermore, we compare our model with this multi-duration model and a multi-duration baseline. Our model improves the saliency prediction in two durations consisting of 0.5 and 3 seconds in two out of three metrics and in all three metrics in the five second duration.

\subsection{Comparison with state-of-the-art methods}
We first evaluate the performance of our model on image saliency prediction on the SALICON benchmark \cite{salicon}. The ground truth of SALICON’s test set is exclusively hosted on the  CodaLab website\footnote{https://competitions.codalab.org/competitions/17136}. Table~\ref{tab:sota} shows the comparison of standard evaluation metrics for different state-of-the art saliency models alongside our model TempSAL. TempSAL outperforms all the baselines in almost all metrics. When it does not, it still yields competitive results.

\begin{table*}[h]
  \centering
\setlength{\tabcolsep}{9pt}
\begin{tabular}{lccccccccc}
\multicolumn{1}{l}{ Model }& MD  & AUC $\uparrow$ & CC $\uparrow$ & KL $\downarrow$ & SAUC $\uparrow$ & IG $\uparrow$ & NSS $\uparrow$ & SIM $\uparrow$ \\
\cmidrule(lr){1-1}\cmidrule(lr){2-2}\cmidrule(lr){3-9}SAM-Resnet ~\cite{sam} & \xmark & $0.865$ & $0.899$ & $0.610$ & $0.741$ & $0.538$ & $1.990$ & $0.793$  \\
MSI-Net ~\cite{KRONER2020261} & \xmark &$0.865$ & $0.899$ & $0.307$ & $0.736$ & $0.793$ & $1.931$ & $0.784$ \\
GazeGAN ~\cite{Che2019GazeGANAG} & \xmark &$0.864$ & $0.879$ & $0.376$ & $0.736$ & $0.720$ & $1.899$ & $0.773$\\
SimpleNet ~\cite{simplenet} & \xmark &$\mathbf{0 . 8 6 9}$ & $0.907$ & $0.201$ & $0.743$ & $0.880$ & $1.960$ & $0.793$ \\
MDNSal ~\cite{simplenet} &\xmark & $0.865$ & $0.899$ & $0.221$ & $0.736$ & $0.863$ & $1.935$ & $0.790$  \\

UNISAL ~\cite{unisal}& \xmark &$0 . 8 6 4$ & $0.879$ & $0.354$ & $0.739$ & $0.780$ & $1.952$ & $0.775$ \\
DeepGaze IIE ~\cite{linardos} & \xmark &$\mathbf{0.869}$ & $0.872$ & $0.285$ & $\mathbf{0.767}$ & $0.766$ & $1.996$ & $0.733$ \\
MD-SEM ~\cite{fosco2020howmuch} & \cmark &$0.864$ & $0.868$ & $0.568$ & $0 . 7 4 6$ & $0.660$ & $\mathbf{2 . 0 5 8}$ & $0.774$  \\
\textbf{TempSAL} & \cmark & $\mathbf{0 . 8 6 9}$ & $\mathbf{0 . 9 1 1}$ & $\mathbf{0 . 1 9 5}$ & ${0.745}$ & $\mathbf{0 . 8 9 6}$ & $1.967$ & $\mathbf{0 . 8 0 0}$ 


\end{tabular}
\caption{Evaluation results on the SALICON (LSUN 2017)
test benchmark. We compare our model with the state-of-the-art saliency prediction models, namely SAM-Resnet ~\cite{sam}, MSI-Net ~\cite{KRONER2020261}, GazeGAN ~\cite{Che2019GazeGANAG}, MDNSal ~\cite{simplenet}, SimpleNet ~\cite{simplenet}, DeepGaze IIE ~\cite{linardos}, and MD-SEM ~\cite{fosco2020howmuch}. The results in bold show the best performance. Our method outperforms the state-of-the-art on conventional image saliency in five metrics. The MD column denotes the ability of the models to predict multi-duration saliency. Our model outperforms the only other multi-duration saliency model by a significant margin on five out of seven metrics. }
\label{tab:sota}
\end{table*}
\begin{table*}[h]
\centering
\setlength{\tabcolsep}{5pt}
\begin{tabular}{lcccccccccccc}

&\multicolumn{3}{c} {Slice 1 (0-500 ms) } &\multicolumn{3}{c} {Slice 2 (0-3000 ms)} &\multicolumn{3}{c} {Slice 3 (0-5000 ms)}&\multicolumn{3}{c} {Average }\\

 Model  & { CC $\uparrow$} & { KL $\downarrow$} & { NSS $\uparrow$} & { CC $\uparrow$} & { KL $\downarrow$} & { NSS $\uparrow$} & { CC $\uparrow$} & { KL $\downarrow$} & { NSS $\uparrow$} & { CC $\uparrow$} & { KL $\downarrow$} & { NSS $\uparrow$} \\
\cmidrule(lr){1-1}\cmidrule(lr){2-4}\cmidrule(lr){5-7}\cmidrule(lr){8-10}\cmidrule(lr){11-13}
 SAM-MD~\cite{fosco2020howmuch}&  $0.805$ & $0.370$ & $ 3.181$  
 & $0.738$ & $0.469$ & $ 2.541$ &
 $0.715$ & $0.535$ & $ 2.495$ & 
 $0.753$ & $0.458$ & $ 2.739$ \\
		 MD-SEM~\cite{fosco2020howmuch} & $0.816$ & \textbf{0.351} & $3.374$  
		 &    $0.745$ & $\textbf{ 0.452}$ & $2.694$       
		 &$0.734$  & $0.487$  & $2.677$           
		 &  $0.765$    & $\textbf{0.430}$ & $2.915$   \\
		\textbf{TempSAL} & $\textbf{0.819}$  & $0.496$  & $ \textbf{3.422} $  
		& $\textbf{0.752}$   & $0.512$  & $\textbf{2.703}$&               
		$\textbf{0.822}$  & $\textbf{0.471}$  & $\textbf{3.337}$  
		& \textbf{0.797}	& $0.493$ & $\textbf{3.154}$

\end{tabular}

 \caption{Results of our model, MD-SEM~\cite{fosco2020howmuch}, and the baseline SAM-MD~\cite{fosco2020howmuch}
 across different durations on the CodeCharts1k dataset~\cite{fosco2020howmuch}. Our model improves the saliency prediction in the first two intervals, consisting of 500 ms and 3000 ms observations, in two out of three metrics. In this comparison, the time slices are cumulative, not mutually exclusive. Although our model benefits from non-overlapping temporal slices, it also performs well with cumulative time slices, particularly in the last slice corresponding to the image saliency with an observation duration of 5000 ms.}  
\label{mdcomparison}
\end{table*}

\subsection{Comparison with the multi-duration method}\label{sec:multiresults}
To compare our method with the only other multi-duration model~\cite{fosco2020howmuch}, we modify our network to output three temporal slices. We train our network on a three slice SALICON multi-duration dataset first and then fine-tune it on the CodeCharts1k dataset~\cite{fosco2020howmuch} using the given training and validation splits. We report the results of the comparison in Table~\ref{mdcomparison}.

\subsection{Qualitative Results}
\begin{figure*}[h]
    \centering
    \includegraphics[width=0.95\textwidth]{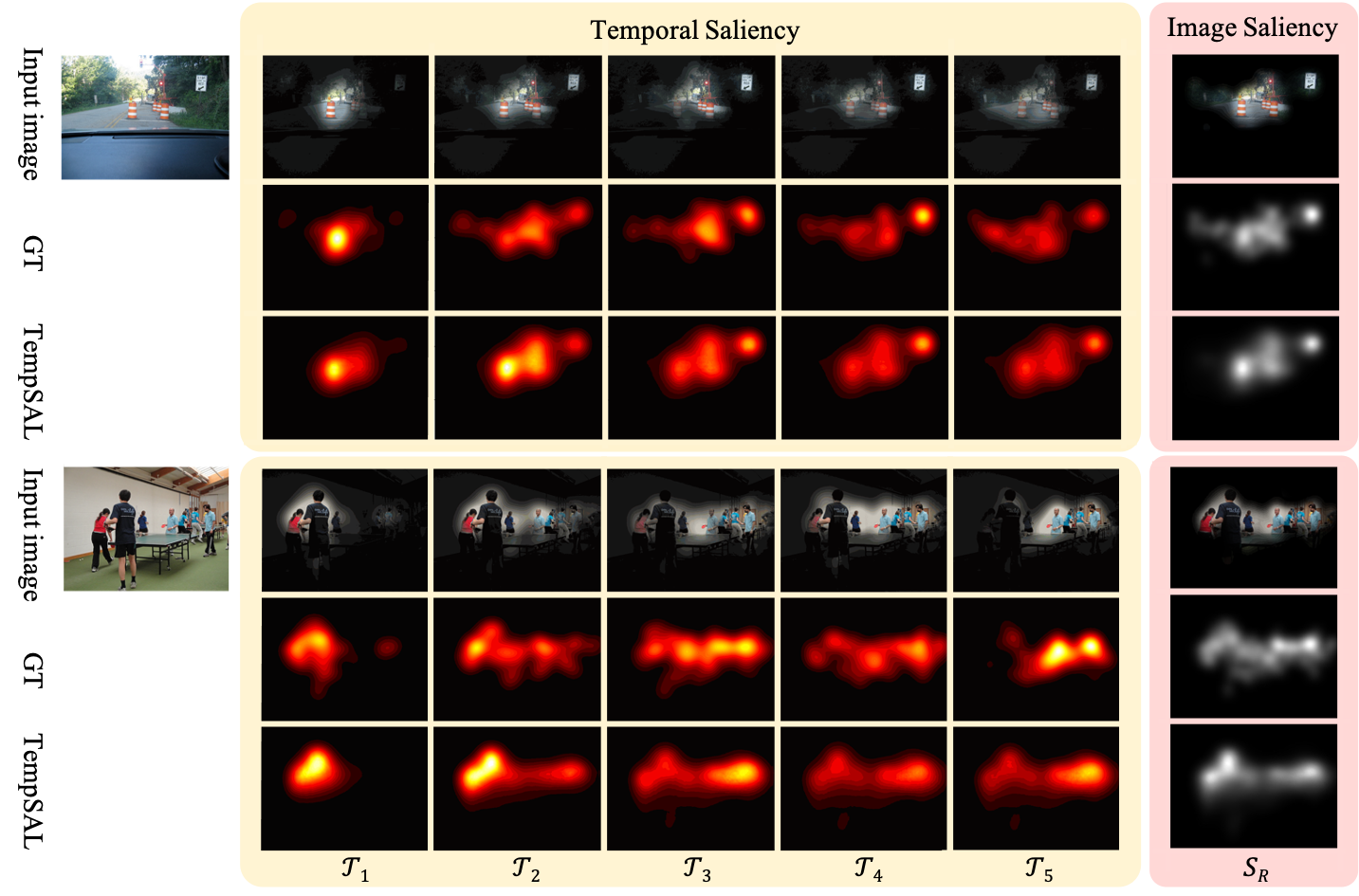}
        \vspace{-8pt}
    \caption{The first row shows the input image with the ground-truth saliency overlaid. The second row shows the ground truth \textbf{\textcolor{myorange}{temporal saliency}}  and \textbf{\textcolor{mypink}{image saliency}}. We show our \textbf{\textcolor{myorange}{temporal saliency}} and \textbf{\textcolor{mypink}{image saliency}} predictions on the  third row. Red-yellow maps are temporal saliency maps for one-second intervals. Black and white maps are image saliency maps for the whole observation duration. Our approach captures the attention shifts in sequential temporal maps. Moreover, our model is able to produce accurate image saliency predictions which are close to the ground truth maps. 
     }
    \label{fig:temporalfigure}
    \vspace{-12pt}
\end{figure*}

In Figure~\ref{fig:temporalfigure}, we compare the temporal and image saliency maps obtained with our method with the ground truth from SALICON~\cite{salicon}. Our model learns time-specific predictions and is  able to combine such predictions with a conventional image saliency map. We provide additional qualitative results in the supplementary material.

\subsection{Ablation studies}
In this section, we investigate the effect of different components in our model, and of two temporal slicing methods. We also provide a comparison with a multi-duration baseline model on the temporal SALICON dataset. Moreover, we provide an ablation study on the effect of the number of time slices and the time-slicing alternatives in the supplementary material.

\textbf{Effect of the SMM module:} We evaluate the effect of the spatiotemporal mixing module (SMM) and the image saliency decoder in Table~\ref{ablationsmm}. The first model consists of the image encoder and temporal saliency decoder only. We take the average of the temporal slices to measure its performance by comparing with the ground-truth image saliency map. In the second row, we add the image saliency decoder to our model. Similarly, we take the average of the predicted maps $\mathcal{T}_{n}$ and $\mathcal{S}_{I}$. Lastly, we add the spatiotemporal mixing module, which effectively modulates these predicted maps and combines them into a final image saliency map $\mathcal{S}_{R}$. 
\begin{table}[h]
\setlength{\tabcolsep}{7pt}

\centering
\begin{tabular}{lcccc}
 Model & { CC $\uparrow$} & { KL $\downarrow$} & { NSS $\uparrow$} & { SIM $\uparrow$} \\
\cmidrule(lr){1-1}\cmidrule(lr){2-5}$\mathcal{D}_{T}(\mathcal{E(I)})$ & $0.852$  & $0.243$  & $\textbf{1.973}$  & $0.754$   \\
  $+ \mathcal{D}_{S}(\mathcal{E(I)})$ & $0.857$  & $0.252$  & $1.943$  & $0.760$   \\
 $+ \mathrm{SMM}$  & $\textbf{0.906}$  & $\textbf{0.198}$  & $1.930$  & $\textbf{0.798}$   \\

\end{tabular}
  \vspace{-5pt}
 \caption{Results of ablation studies on the temporal SALICON validation dataset. The first row denotes the model with only the temporal saliency decoder. In the second row, the model has both the temporal and image saliency decoders. The last row denotes the performance with the spatiotemporal mixing module (SMM). As evidenced by the improved accuracy metrics, the SMM effectively modulates the spatial and temporal saliency maps to refine the initial image saliency prediction. }
 \label{ablationsmm}
  \vspace{-15pt}
\end{table}

\textbf{Comparison with a temporal baseline model:}\mbox{}
The performance of our TempSAL model with five temporal slices is provided in Table~\ref{tab:ourbaseline}. Note that each saliency slice contains five times fewer samples than the original image saliency map. Therefore, individual slices contain more variation compared to conventional accumulated maps. 
As a baseline to our model, we compute the performance of an architecture composed of five replicated SimpleNet models (5xSimpleNet)~\cite{simplenet}, each trained on one saliency slice. This baseline model uses an unshared encoder and decoder for each slice, while we share the decoder among slices. Therefore, we do not benefit from increased model capacity. We observe an accuracy decline in the baseline model, which confirms the increased discrepancy in the data.
\begin{table}[h]
    \vspace{-10pt}
\setlength{\tabcolsep}{2pt}
\begin{center}
\resizebox{\columnwidth}{!}{
\begin{tabular}{ccccccccc}
 &\multicolumn{4}{c} {Baseline} &\multicolumn{4}{c} {\textbf{TempSAL}}\\
 Time & { CC $\uparrow$} & { KL $\downarrow$} & { NSS $\uparrow$} & { SIM $\uparrow$} & { CC $\uparrow$} & { KL $\downarrow$} & { NSS $\uparrow$} & { SIM $\uparrow$}\\
\cmidrule(lr){1-1}\cmidrule(lr){2-5}\cmidrule(lr){6-9}
$\mathcal{T}_{1}$& $0.898$  & $0.211$  & $2.436$  & $0.778$		    & $0.899$ & $0.214$ & $2.453$ & $0.782$  \\
		$\mathcal{T}_{2}$& $0.870$  & $0.219$  & $2.159$  & $0.765$ 			& $0.877$ & $0.215$ & $2.211$ & $0.776$  \\
		$\mathcal{T}_{3}$ & $0.840$  & $0.247$  & $1.840$  & $0.753$		    & $0.843$ & $0.247$ & $1.878$ & $0.758$  \\
		$\mathcal{T}_{4}$& $0.820$  & $0.273$  & $1.729$  & $0.743$ 			& $0.825$ & $0.264$ & $1.740$ & $0.749$  \\
		$\mathcal{T}_{5}$ & $0.811$  & $0.275$  & $1.646$  & $0.738$ 			& $0.813$ & $0.276$ & $1.654$ & $0.743$ \\
\rowcolor{mygray}
Average & $0.848$  & $0.245$ & $1.962$ & $0.756$ 			& $\mathbf{0 . 8 5 2}$&  $\mathbf{0 . 2 4 3}$ & $\mathbf{1 . 9 8 7}$   & $\mathbf{0 . 7 6 1}$
\end{tabular}
}
  \vspace{-5pt}
\caption{Results of the baseline model (left) and our TempSAL model (right) across different time slices. In 18 out of 20 comparisons, our model consistently outperforms the baseline. Note that both models perform best in the first slice, in which the intra-slice agreement is more prominent than in the other slices, as mentioned in Section~\ref{sec:temporal}. }
\label{tab:ourbaseline}
\vspace{-30pt}
\end{center}
\end{table}

\section{Conclusion}
We present a saliency prediction method that can learn time-specific predictions and is also able to exploit temporal information to improve overall image saliency prediction. In particular, we show that the temporally evolving patterns in human attention play an important role in saliency prediction in natural images. This is evidenced by our experiments that demonstrate our TempSAL method outperforming the state-of-the-art, including a multi-duration method exploiting cumulative temporal saliency maps.\\

\vspace{-9pt}
\textbf{Acknowledgement.} This work was supported by the Swiss National Science Foundation via the Sinergia grant CRSII5-180359.

\clearpage

%
%

{\small
\bibliographystyle{ieee_fullname}
\bibliography{mybib}
}
\clearpage

\end{document}